\documentclass{article}
 \usepackage{graphicx}
\usepackage{amsthm}
\usepackage{amsmath}
\usepackage{amssymb}
\usepackage{hyperref}

\newcommand{\asin}{\mbox{asin}}
\newcommand{\acos}{\mbox{acos}}
\newcommand{\atan}{\mbox{atan}}
\newcommand{\acot}{\mbox{acot}}
\newcommand{\acsc}{\mbox{acsc}}
\newcommand{\asec}{\mbox{asec}}
\newcommand{\csch}{\mbox{acsch}}
\newcommand{\sech}{\mbox{asech}}
\newcommand{\asinh}{\mbox{asinh}}
\newcommand{\acosh}{\mbox{acosh}}
\newcommand{\atanh}{\mbox{atanh}}
\newcommand{\acoth}{\mbox{acoth}}
\newcommand{\acsch}{\mbox{acsch}}
\newcommand{\asech}{\mbox{asech}}

\newcommand{\goesto}{\mbox{$\rightarrow$}}

\title{A Flawed Dataset for Symbolic Equation Verification}

\author{
Ernest Davis \\
Dept. of Computer Science \\ New York University \\ New York, NY 10012 \\
{\small davise@cs.nyu.edu}}

\begin{document}
\maketitle

\begin{abstract}
Arabshahi, Singh, and Anandkumar (2018) propose a method for creating a dataset
of symbolic mathematical equations for the tasks of symbolic equation verification
and equation completion.  Unfortunately, a dataset constructed using the method
they propose will suffer from two serious flaws. First, the
class of true equations that the procedure can generate will be very limited.
Second, because true and false equations are generated in completely different
ways, there are likely to be artifactual features that allow easy discrimination.

Moreover, over the class of equations they consider, 
there is an extremely simple probabilistic procedure that solves
the problem of equation verification with extremely high reliability.
The usefulness of this problem in general as a testbed for AI systems
is therefore doubtful.
\end{abstract}

\section{Introduction}

Arabshahi, Singh, and Anandkumar (2018) (henceforth ASA) 
propose a method for creating datasets
of symbolic mathematical equations for the tasks of symbolic equation verification
and equation completion.  At least one other research group has used their
method of dataset construction to continue this line of research (Mali et al.
2021).

There are two tasks, with two related but different datasets, addressed in
ASA.
The first task, which I will discuss in this review, 
is {\em equation verification}: given
a syntactically correct formula, determine whether or not it is true. For 
instance, the equations 2+2=4, $\sin^{2}(x)+\cos^{2}(x)=1$, and
$(x+y)^{2} = x^{2}+2xy+y^{2}$  are true equations;
2+2=3, $\sin(\sin(x^{2})) = \cos(x)$, and $x+y^{2}=1$ are false equations. 

The second task is the task of {\em equation completion\/}; given an equation
from which a symbol has been omitted, find the missing symbol. I will
not discuss that task further in this review. Suffice it to say that
it is not a mathematically meaningful or useful task. It seems likely that
it was chosen because it corresponds to the natural abilities of transformer
technology.

Unfortunately, the dataset constructed for equation verification in ASA
suffers from flaws of two kinds:
\begin{itemize}
\item The class of true equations that the 
procedure can generate will be very limited.
\item Because true and false equations are generated in significantly different
ways, there are likely to be artifactual features that often allow easy 
discrimination.
\end{itemize}
Moreover, over the class of equations they consider, 
there is an extremely simple probabilistic procedure that solves
the problem of equation verification with extremely high reliability.
The usefulness of this problem in general as a testbed for AI systems
is therefore doubtful.

\section{How the dataset is constructed}
The paper (Arabshahi, Singh, and Anandkumar, 2018) describes the language of
equtions as follows: (As we will see in section~\ref{secDiscrepancies}, 
the language actually used in the set of axioms
is significantly different.) Equations
are symbolic trees in a context-free
grammar. The root is always the equality sign `='. The two children of the
root are terms. The leaves of the tree are the variable symbols `x', `y',
and `z'; the numbers
$-1$, 0, 1, 2, 3, 4,  10 and Half; and numbers with two digits
of precision between $-3.14$ and 3.14, such as 2.2 or $-1.5$.
The internal nodes other than the root are 
arithmetic, binary exponential, $e^{x}$,
trigonometric, inverse trigonometric,
hyperbolic, and inverse hyperbolic functions 
(table~\ref{tabSymbols}).

\begin{table}
{\bf Terminals:} $-1$, 0, 1, 2, 3, 4, 10, $\pi$, $x,y,z$, \\
\hspace*{3em} and real numbers 
with two digits of precision between $-3.14$ and 3.14. \\
{\bf Binary functions:} Addition, multiplication, exponentiation. \\
{\bf Unary functions:} sin, cos, sec, tan,
cot, arcsin, arccos, arccsc, arcsec, arctan, arccot, sinh, 
cosh, scsh, sech, tanh, coth, arsinh, arsech, artanh, arcoth, exp.
\caption{Symbols in the language, as described in the paper}
\label{tabSymbols}
\end{table}

The starting point for each equation, both correct and incorrect, is a collection
of about 140 identities, 66 arithmetic and the rest trigonometric. No
valid axioms are given for the hyperbolic functions.
Each identity can be considered as a transformation
rule; for instance the identity $x+y=y+x$ can be considered the transformation,
``Replace $x+y$ by $y+x$.''

To generate a false equation, the system starts with a true equation and then 
randomly modifies it, using a number of specified categories of transformations.
With near certainty, the result will be a false statement; to 
make sure of this, they ran the result through the external system {\em 
sympy} (Meurer et al. 2017). For instance, one transformation is to replace a
symbol by a randomly chosen symbol with the same syntactic role.
If you start with the equation $\cos^{2}(z)+\sin^{2}(z) =1$
and replace the first occurrence of 2 by 3, you get the false equation
$\cos^{3}(z)+\sin^{2}(z) =1$.

To generate a true equation, the system starts with a true equation and performs
substitutions on either side that are supported by the transformation rules.
For instance, if you start with the equation $\cos^{2}(\theta)+\sin^{2}(\theta) =1$
and apply the transformation rule $x+y \goesto y+x$ to the left-hand
side, you get the new equation
$\sin^{2}(\theta)+\cos^{2}(\theta) =1$

\section{Flaws in the dataset}
This method of constructing the dataset has two kinds of inherent flaws. First,
the coverage of true equations is woefully inadequate; many simple true equations
will never be generated. Second, there are likely to be systematic artifactual
differences between the true and the false equations which can be 
easily detected. Moreover, the set of primitives in the axioms 
differs significantly from the set of primitives described in the paper.

\subsection{Inadequate coverage of true equations}
\label{secInadequate}
There are numerous simple true equations that are in the ASA language but which
the method cannot construct, for two reasons.

First, the system can only substitute for one term at a time in an equation; it
cannot do substitutions that involve repeated occurrences of a variable.
Thus, for instance, starting with the equation $x+y=y+x$, it cannot substitute
2 for $x$ and $\pi$ for $y$ and derive the equation $2+ \pi = \pi+2$. In fact,
it has no path to derive that equation through the process that generates
true equation, though the process that generates false equations might
derive it, purely by chance, with very low probability. (If it was so
derived, then it would be run through sympy, and would be marked as true.)

Second, the list of arithmetic axioms, shown in the appendix below, 
is altogether {\em ad hoc} and is
inadequate to the language and the task.
For example, the distributive law $x * (y + z) = x * y + x * z$ and
the laws of exponents $x^{y} \cdot x^{z} == x^{y+z}$  and
$(x^{y})^{z} = x^{y \cdot z}$ are missing.
On the other hand, it includes
11 pure identities ($x+x==x+x$, $x=x$, and similarly for 1, 0, 2, 4, 3, 
Half, Pi, 10, and $-1$); I do not understand what function these can possibly
serve.  It has a large number of equations with constants, many of which
are special cases of more general rules, such as $-1 * 0 = 0$ and 
$0 * -1 = 0$, but by no means a complete set over the constants in the
language.

These problems could be alleviated by using a more general sense of
substitution and a better chosen set of axioms.

\subsection{Artifactual differences}
True and false equations are constructed in very different ways. As a result,
there are simple, easily detected, patterns that are much more likely to 
occur in false equations than in true ones and vice versa.

For example, an equation that contains the direct composition of two transcendental
function or a transcendental function to a non-integer power is almost certainly
false, particularly if the term appears only once in the equation. If an
equation includes $\sin(\sin(x))$ or $\sin(x)^{y+2}$, it is a pretty good bet
that it is false.  There are exceptions, certainly: 
$\sin^{2}(\sin(x))+\cos^{2}(\sin(x))=1$ 
and $0 \cdot \sin(\sin(x))= 0$
are true equations. However, ASA
cannot construct them, for the reasons discussed in section~\ref{secInadequate}.
My guess is that the ASA procedure for constructing true equations
cannot construct {\em any\/} true equation that includes the
term $\sin(\sin(x))$, though I could be mistaken.
In the other direction, if an equation contains a complex expression strongly
associated with one of the standard identities, not at the top level of the 
equation, then it is a pretty good guess that it was 
constructed by the procedure for constructing true equations. 
For instance, if you see the equation 
$\sin^{2}(x)+\cos^{2}(x)+2=3$, it is a safe bet that the true equation
generator started with $1+2=3$ and then substituted 
$\sin^{2}(x)+\cos^{2}(x)$ for 1. It is {\em possible\/} that the false
equation generator might stumble across that expression purely at random,
and generate a false equation like
$\sin^{2}(x)+\cos^{2}(x)+2=4$, but it is very unlikely.
The false equation generator will, with some reasonable frequency, 
start with the equation $\sin^{2}(x)+\cos^{2}(x)=1$ and then only modify
the right-hand side, leading to false equations like
$\sin^{2}(x)+\cos^{2}(x)=4$, so this heuristic only applies strongly when
the expression is not the whole of one side of the equation.

The second problem can be mitigated or eliminated by having the false equation
generator apply many valid transformations and only one invalid transformation.
The first problem can be mitigated, either crudely, by applying a filter
for these kinds of improbable expressions, or, with greater complexity,
by limiting the kinds of transformations that the false equation generator
can use to ones that look plausible. Both problems would be significantly
mitigated if the true equation generator were made more powerful.

\subsection{Discrepancies in the language between the paper and the
axiom set}
\label{secDiscrepancies}
The primitives in the language as described in ASA differ substantially
from those used in the axiom sets associated with the paper on github.

The axioms include the symbols Half, minus, and divide, which are not
in the paper's description of the language.
ASA state: ``Note that we exclude subtraction and division because they
can be represented with addition, multiplication, and exponentiation
respectively [sic].'' However, that is not the representation used in the
axioms, and, of course, in this kind of symbolic manipulation system,
the specifics of the representation matter. 

The first trigonometric axiom is a very long, nonsensical
entirely false, equation, as follows: 

\begin{quote}
$\sin(\theta)+\cos^{2}(\theta)+\sin(\pi)+\frac{1}{2}- 
\frac{1}{2}\cos(2 \theta) + \sec(\theta) \csc(\theta) =$ \\
$\frac{1}{\csc(\theta)} + \tan(\theta) - \cot(\theta) + 
\log(\asin(\theta)) - \mbox{exp}(\acos(\theta)) + \asec(\theta) +$ \\
$\acsc(\theta) + \sqrt{\atan(\theta)} + \acot(\theta) + 1 +
\sinh(\theta) + \cosh(\theta) - \coth(\theta) + \tanh(\theta)$ \\
$ \csch(\theta) + \sech(\theta) + \acsch(\theta) + \asech(\theta) +$ \\
$+ \asinh(\theta) + \acosh(\theta) + \acoth(\theta) + \atanh(\theta) + 
\sqrt{\theta}$  
\end{quote}

How or why this got included is not at all clear. This axioms includes
the functions $\sqrt{x}$ and log which are not in the paper's description
of the language. If one excludes this axiom as nonsensical, then the hyperbolic
functions and the exponential function do not appear in the axiom set. 

It does not affect the system, but it is characteristic of this project
that the English description of the trigonometric axioms in the collection
trigonometric axioms has been carelessly cut and
pasted; axioms 1-21 are {\em all\/} labelled ``The tangent of
an angle is the ratio of the sine to the cosine.''

\section{A simple method for equation verification}
\label{secSimpleMethod}
In the worst case, equation verification over the language of ASA is 
undecidable (Richardson, 1969). But those undecidable cases are exceedingly
rare. In fact, there is a simple probabilistic 
method for equation verification that is extremely effective almost all
of the time (Davis, 1977; Schwartz, 1980).

Step 0: Let $\epsilon > 0$ be a small number. \\ 
Step 1: Choose random values for the variables. \\
Step 2: Evaluate the left- and right-hand sides of the equation using floating
point arithmetic in a high-quality system. If they are within $\epsilon$, then the
equation is almost certainly true, if they are more than $\epsilon$ apart
then then the equation is almost certainly false.

If you need more certainty, in equations that have variables, then iterate
with more random values. If you ever get values that are far apart, then the
equation is almost certainly false.

For instance, if we want to verify the equation 
$\sin^{2}(\sin(x))+\cos^{2}(\sin(x))=1$, we can randomly choose $x=1.6$, say. 
Using double-precision arithmetic in Matlab, the left side of the equation evaluates
to 1 with no error.
If we want to disconfirm the equation
$\sin^{2}(\sin(x))+\cos^{3}(\sin(x))=1$, we can again choose $x=1.6$. In
this case the left hand side evaluates to 0.8657, so as long as 
$\epsilon < 0.13$ we can rule this out.

Occasionally, this method can go wrong, in both directions, 
due to round-off error. 
Applying the method using Matlab to the 
equations $(x+y)-x=0$ and $(x+y)-x=y$,
it will say that the first is true and the second is false, if you happen to choose
$x=10^{9}$ and $y=10^{-9}$. Or it can go into overflow or underflow, as
will happen if you try to numerically verify 
$\sin^{2}(10^{10^{10}})+
\cos^{2}(10^{10^{10}}) = 1$
There are also occasionally 
near identities, such as Ramanujan's approximation $22 \pi^{4} \approx 2143$, 
which is accurate to within $3 \cdot 10^{-6}$.
But that will be very rare, over the kind of equations
that ASA generates. (As the complexity of the expressions
increases, the likelihood that a 
random equation can be reliably validated using floating point evaluation
goes to zero, but presumably, over the length of expressions that ASA 
consider, it remains close to 1.)

Does this inherent simplicity matter? Well, it certainly somewhat deflates the
claim that state-of-the-art AI achieves 98\% accuracy when you realize that 
in a few lines of code, you can write a program that achieves 99.9999\% 
(at a guess) accuracy. (Of course, the few lines of code do have to be
supplemented by a high-quality mathematical library for computing the
functions in floating point.)
It seems unlikely that the deep learning systems 
that have been tested on this problem have in any sense taken advantage of this,
but it would be difficult to be sure. 

\section{Conclusion}
If the goal in constructing such a dataset is merely to provide a standard on
in which various deep learning architectures can compete, without reference 
to the true underlying task, then perhaps these flaws do not matter. 
Nonetheless, 
for the health of the field in general, it is important that scientists should
be aware of them. If, based on this data set,  
it is reported that state-of-the-art machine learning
technology can achieve 98.8\% accuracy over the problem of 
equation verification
(Mali et al., 2021)
then that naturally conveys the impression that 
great progress is being made in  
symbolic reasoning. In view of the profound flaws in the dataset, it actually
indicates next to nothing. 

All in all, the uselessness of the equation completion task, the
shoddiness of the set of algebraic axioms, 
the inadequate mechanisms for generating the data set,
and the discrepancies between the actual language used and its description
in the paper
suggest strongly that the authors
have no serious interest in symbolic mathematics. It seems to
me that a scientist who is applying machine learning techniques to
a domain has a responsibility to take that domain somewhat seriously.

% \subsection*{Acknowledgements}

\subsection*{References}
Arabshahi, Forough, Sameer Singh, and Animashree Anandkumar (2018). 
``Combining symbolic expressions and black-box function evaluations in 
neural programs.'' {\em ICLR-2018.} 
\url{https://arxiv.org/abs/1801.04342} 

Davis, Philip (1977). ``Proof, Completeness, Transcendentals, 
and Sampling.'' {\em J. ACM.} {\bf 24}(2): 298-310.

Mali, Ankur, Alexander Ororbia, Daniel Kifer, and C. Lee Giles (2021). 
``Recognizing and Verifying Mathematical Equations using Multiplicative 
Differential Neural Units.'' arXiv preprint arXiv:2104.02899 (2021).
\url{https://arxiv.org/abs/2104.02899}

Meurer, Aaron et al. (2017). Sympy: Symbolic computing in Python.
{\em PeerJ Computer Science,} e103. 
\url{https://peerj.com/articles/cs-103.pdf}

Richardson, Daniel (1969). ``Some undecidable problems involving elementary 
functions of a real variable." 
{\em The Journal of Symbolic Logic,} {\bf 33}:4, 514-520.

Schwartz, Jacob (1980). ``Fast probabilistic algorithm for verification
of polynomial identities.'' {\em J. ACM}, {\bf 27}(4): 701-717.

\subsection*{Appendix: The basic algebraic equations used in ASA}
From
\url{https://github.com/ForoughA/neuralMath/tree/master/axioms}

\begin{verbatim}
  1.  1 + 1 == 2
  2.  1 * 1 == 1
  3.  0 * 1 == 0
  4.  0 * 2 == 0
  5.  0 * 3 == 0
  6.  0 * 0 == 0
  7.  1 ** Half == 1
  8.  1 ** x == 1
  9.  x ** 1 == x
 10.  x ** 0 == 1
 11.  4 ** Half == 2
 12.  1 ** Half == 1
 13.  2 ** 2 == 4
 14.  1 ** 1 == 1
 15.  2 ** 0 == 1
 16.  3 ** 0 == 1
 17.  4 ** 0 == 1
 18.  4 ** Half == 2  [sic -- this is repeated from 11]
 19.  3 + 1 == 4
 20.  (1 + 1) + 1 == 3
 21.  2 + 1 == 3
 22.  -1 + 3 == 2
 23.  3 - 1 == 2
 24.  4 - 1 == 2
 25.  1 - 1 == 0
 26.  0 - 0 == 0
 27.  0 + 0 == 0
 28.  0 + x == x
 29.  0 * x == 0
 30.  1 * x == x
 31.  x + x ==  2 * x
 32. (x + x) + x  == 3 * x
 33. x + x == x + x 
 34. x + y == y + x
 35. (x + y) + z == x + (y + z)
 36. (x + z) + y == x + (y + z)
 37. (z + x) + y == y + (z + x)
 38. (z + x) + y == y + (x + z)
 39. (x + y) + z == x + (y + z)
 40. (x + y) + z == x + (y + z)
 41. 1 * (x + y) == x + y
 42. 1 * (x + y) == 1 * x + 1 * y
 43. 2 * (x + y) == 2 * x + 2 * y
 44. x * y == y * x
 45. (x * y) * z == x * (y * z)
 46. (x * z) * y == x * (y * z)
 47. (z * x) * y == x * (y * z)
 48. (z * x) * y == y * (z * x)
 49. (x * y) * z == y * (x * z)
 50. (x * y) * z == y * (z * x)
 51. -1 * 0 == 0
 52. 0 * -1 == 0
 53. -1 * 1 == -1
 54. 1 * -1 == -1
 55. -1 * 2 == -2
 56. 2 * -1 == -2
 57. x == x
 58. 1 == 1
 59. 0 == 0
 60. 2 == 2
 61. 3 == 3
 62. 4 == 4
 63. Half == Half
 64. Pi == Pi
 65. 10 == 10
 66. 1 == 1
\end{verbatim}

\end{document}